\pdfoutput=1

\documentclass[11pt]{article}
\usepackage{authblk}

\usepackage{emnlp2021}

\usepackage{times}
\usepackage{latexsym}
\usepackage[T1]{fontenc}
\usepackage[utf8]{inputenc}
\usepackage{microtype}

\usepackage{amsmath}
\usepackage{multirow}
\usepackage{xcolor}
\usepackage{dashbox}

\usepackage{fbox}
\usepackage{longfbox}

\usepackage{soul}
\usepackage{booktabs}
\usepackage{xstring}
\usepackage{fontawesome}
\usepackage{tikz-dependency}
\usepackage{varwidth}
\usepackage{gb4e}
\usepackage{tabularx}

\noautomath

\newcommand{\cfbox}[2]{%
    \colorlet{currentcolor}{.}%
    {\color{#1}%
    \framebox{\color{currentcolor}#2}}%
}

\newcommand*{\email}[1]{%
    \small\href{mailto:#1}{#1}\par
    }
\title{Anatomy of OntoGUM---Adapting GUM to the OntoNotes Scheme to Evaluate Robustness of SOTA Coreference Algorithms}

\author[1]{\textbf{Yilun Zhu}}
\author[2,3]{\textbf{Sameer Pradhan}}
\author[1]{\textbf{Amir Zeldes}}
\affil[1]{Department of Linguistics, Georgetown University}
\affil[2]{Linguistic Data Consortium, University of Pennsylvania}
\affil[3]{cemantix.org}

\affil[ ]{\tt \email{yz565@georgetown.edu, pradhan@cemantix.org, Amir.Zeldes@georgetown.edu}}

\begin{document}
\maketitle

\begin{abstract}
SOTA coreference resolution produces increasingly impressive scores on the OntoNotes benchmark. 
However lack of comparable data following the same scheme for more genres makes it difficult to evaluate generalizability to open domain data. 
\citet{zhu-etal-2021-ontogum} introduced the creation of the OntoGUM corpus for evaluating geralizability of the latest neural LM-based end-to-end systems.  This paper covers details of the mapping process which is a set of deterministic rules applied to the rich syntactic and discourse annotations manually annotated in the GUM corpus. 
Out-of-domain evaluation across 12 genres shows nearly 15-20\% degradation for both deterministic and deep learning systems, indicating a lack of generalizability or covert overfitting in existing coreference resolution models. 
\end{abstract}

\section{Introduction}
Coreference resolution is the task of grouping referring expressions that point to the same entity, such as noun phrases and the pronouns that refer to them. The task entails detecting correct mention or `markable' boundaries and creating a link with previous mentions, or antecedents. A coreference chain is a series of decisions which groups the markables into clusters. As a key component in Natural Language Understanding (NLU), the task can benefit a series of downstream applications such as Entity Linking, Dialogue Systems, Machine Translation, Summarization, and more \cite{poesio2016anaphora}.

In recent years, deep learning models have achieved high scores in coreference resolution. The end-to-end approach \cite{lee-etal-2017-end, lee-etal-2018-higher} jointly scoring mention detection and coreference resolution currently not only beats earlier rule-based and statistical methods but also outperforms other deep learning approaches \cite{wiseman-etal-2016-learning, clark-manning-2016-deep, clark-manning-2016-improving}. Additionally, language models trained on billions of words significantly improve performance by providing rich word and context-level information for classifiers \citep{lee-etal-2018-higher, DBLP:journals/corr/abs-1907-10529, Joshi2019BERTFC}.



\newcolumntype{C}[1]{>{\centering\let\newline\\\arraybackslash\hspace{0pt}}m{#1}}

\begin{table*}[t!hb]
    \centering \small
    \begin{tabular}{l|C{1.45cm}|C{1.45cm}|C{1.45cm}|C{1.45cm}|C{1.45cm}|C{1.45cm}|C{1.45cm}}
    Genre & Documents & Tokens & Mentions & Proper & Pron. & Other & Clusters \\
    \hline\hline
    \textit{academic} & 16 & 15,112 & 1,232 & 283 & 262 & 687 & 421 \\
    \textit{bio} & 20 & 17,963 & 2,312 & 934 & 796 & 582 & 487 \\
    \textit{conversation} & 5 & 5,701 & 1,027 & 40 & 728 & 259 & 176 \\
    \textit{fiction} & 18 & 16,312 & 2,740 & 259 & 1,700 & 781 & 469 \\
    \textit{interview} & 19 & 18,060 & 2,622 & 501 & 1,223 & 898 & 608 \\
    \textit{news} & 21 & 14,094 & 1,803 & 796 & 340 & 667 & 477 \\
    \textit{reddit} & 18 & 16,286 & 2,297 & 117 & 1,336 & 844 & 578 \\
    \textit{speech} & 5 & 4,834 & 601 & 171 & 245 & 185 & 134 \\
    \textit{textbook} & 5 & 5,379 & 466 & 108 & 165 & 193 & 133 \\
    \textit{vlog} & 5 & 5,189 & 882 & 22 & 600 & 260 & 149 \\
    \textit{voyage} & 17 & 14,967 & 1,339 & 564 & 300 & 475 & 348 \\
    \textit{whow} & 19 & 16,927 & 2,057 & 53 & 1,001 & 1,003 & 491 \\
    \hline
    Total & 168 & 150,824 & 19,378 & 3,848 & 8,696 & 68,34 & 4,471 \\
    \hline
    \end{tabular}

    \caption{Genre-breakdown Statistics of OntoGUM.}
    \label{tab:genre_stats}
\end{table*}

However, scores on the identity coreference layer of the benchmark OntoNotes dataset \cite{pradhan-etal-2013-towards} do not reflect the generalizability of these systems. \citet{moosavi-strube-2017-lexical} pointed out that lexicalized coreference resolution models, including neural models using word embeddings, face a covert overfitting problem because of a large overlap between the vocabulary of coreferring mentions in the OntoNotes training and evaluation sets. This suggests that higher scores on OntoNotes-test may not indicate a better solution to the coreference resolution task. 

To investigate the generalization problem of neural models, several projects have tested other datasets consistent with the OntoNotes scheme. \citet{moosavi-strube-2018-using} conducted out-of-domain evaluation on WikiCoref \cite{ghaddar-langlais-2016-wikicoref}, a small dataset employing the same coreference definitions. Results showed that neural models (with fixed embeddings) do not achieve comparable performance (16.8\% degradation in score) to scores on OntoNotes.
More recently, the e2e model using BERT \cite{Joshi2019BERTFC} showed gains on the GAP corpus \cite{webster2018gap} using contextualized embeddings; however GAP only contains name-pronoun coreference, a very specific subset of coreference, and is limited in domain to the same single source -- Wikipedia. 

Though previous work has already identified the overfitting problem, it has three main shortcomings. First, the scale of out-of-domain evaluation has been small and homogeneous: WikiCoref only contains 30 documents with $\sim$60K tokens,  much smaller than the OntoNotes test set, and the single genre Wiki domain in both WikiCoref and GAP is arguably not very far from some OntoNotes materials. 
Second, pretrained LMs, e.g.~BERT \cite{devlin2018bert}, popularized after the WikiCoref paper, can learn better representations of markables and surrounding sentences. Aside from GAP, which targets a highly specific subtask, no study has investigated whether contextualized embeddings encounter the same overfitting problem identified by \citeauthor{moosavi-strube-2017-lexical}. Third, previous work may underestimate the performance degradation on WikiCoref in particular due to bias: In \citet{moosavi-strube-2018-using}, embeddings were also trained on Wikipedia themselves, potentially making it easier for the model to learn coreference relations in Wikipedia text, despite limitations in other genres.


In this paper, we explore the generalizability of existing coreference models on a new benchmark dataset, which we make freely available. Compared with work using WikiCoref and GAP, our contributions can be summarized as follows:


\begin{itemize}
    \item We propose OntoGUM, the largest open, gold standard dataset consistent with OntoNotes, with 168 documents ($\sim$150K tokens, 19,378 mentions, 4,471 coref chains) in 12 genres,\footnote{\textbf{Written:} News/Fiction/Bio/Academic/Forum/Travel/How-to/Textbook; \textbf{Spoken:} Interview/Political/Vlog/Conversation.} including conversational genres, which complement OntoNotes for training and evaluation.
    
    \item We show that the SOTA neural model with contextualized embeddings encounters nearly 15\% performance degradation on OntoGUM, showing that the overfitting problem is not overcome by contextualized language models.
    
    \item We give a genre-by-genre analysis for two popular systems, revealing relative strengths and weaknesses of current approaches and the range of easier/more difficult targets for coreference resolution.
    
    
\end{itemize}

\section{Related Work}





\paragraph{OntoNotes and similar corpora}

OntoNotes is a human-annotated corpus with documents annotated with multiple layers of linguistic information including syntax, propositions, named entities, word senses, and within-document coreference \citep{weischedel-handbook-2011-ontonotes,pradhan-etal-2013-towards}.  It covers three languages---English, Chinese and Arabic. The English subcorpus has 3,493 documents and $\sim$1.6 million words annotated for coreference. WikiCoref, which is annotated for anaphoric relations, has 30 documents from English Wikipedia \cite{ghaddar-langlais-2016-wikicoref}, containing 7,955 mentions in 1,785 chains, following OntoNotes guidelines.

\paragraph{GUM} The Georgetown University Multilayer (GUM) corpus \citep{Zeldes2017} is an open-source corpus of richly annotated texts from 12 genres, including 168 documents and over 150K tokens. Though it originally contains more coreference phenomena than OntoNotes using more exhaustive guidelines, it also contains rich syntactic, semantic and discourse annotations which allow us to create the OntoGUM dataset described below. The syntactic annotations, which consist of manually annotated Universal Dependencies trees \cite{10.1162/coli_a_00402}, are particularly useful in analyzing and converting the GUM corpus. We also note that due to its smaller size (currently about 10\% the size of the OntoNotes coreference dataset), it is not possible to train SOTA neural approaches directly on this dataset while maintaining strong performance.

\paragraph{Other corpora} As mentioned above, GAP is a gender-balanced labeled corpus of ambiguous pronoun-name pairs, used for out-of-domain evaluation but limited in coreferent types and genre. Several other comprehensive coreference datasets exist as well, such as ARRAU \cite{poesio-etal-2018-anaphora} and PreCo \cite{chen-etal-2018-preco}, but these corpora cannot be used for out-of-domain evaluation because they do not follow the OntoNotes scheme. Their conversion has not been attempted to date.

\paragraph{Coreference resolution systems} Prior to the introduction of deep learning systems, the coreference task was approached using deterministic linguistic rules \citep{lee-etal-2013-deterministic, recasens-etal-2013-life} and statistical approaches
\citep{durrett-klein-2013-easy, durrett-klein-2014-joint}.
More recently, three neural models achieved SOTA performance on this task: 1) ranking the candidate mention pairs \citep{wiseman-etal-2015-learning,clark-manning-2016-deep}, 2) modeling global features of entity clusters \citep{clark-manning-2015-entity, clark-manning-2016-improving, wiseman-etal-2016-learning}, and 3) end-to-end (e2e) approaches with joint loss for mention detection and coreferent pair scoring \cite{lee-etal-2017-end, lee-etal-2018-higher, fei-etal-2019-end}. The e2e method has become the dominant one, gaining the best scores on OntoNotes. To investigate differences between deterministic and deep learning models on unseen data, we evaluate the two approaches on OntoGUM.

\section{Dataset Conversion}

GUM's annotation scheme subsumes all markables and coreference chains annotated in OntoNotes, meaning we do not need human annotation to recognize additional mentions in the conversion process, though mention boundaries differ subtly (e.g.~for appositions and verbal mentions). Since GUM has gold syntax trees, we were able to process the entire conversion automatically. Additionally, most coreference evaluations use gold speaker information in OntoNotes, which is available in GUM (for \textit{fiction}, \textit{reddit} and spoken data) and could be assembled automatically as well.

The conversion is divided into two parts: removing coreference relations not included in the OntoNotes scheme, and removing or adjusting markables. For coreference relation deletion, we cut chains by removing expletive cataphora, and identifying the definiteness of nominal markables, since indefinites cannot be anaphors in OntoNotes - this was done based on the lemma of determiners attached to the head token of the span (or lack of a determiner) and the POS tag of the head. In addition to modifying existing mention clusters, we also remove particular coreference relations and mention spans, such as Noun-Noun compounding (only included in OntoNotes for proper-name modifiers), bridging anaphora, copula predicates, nested entities (`i-within-i'= single mentions containing coreferring pronouns), and singletons, i.e.~mentions that are not referred to again (all not included in OntoNotes). We note that singletons are removed as the final step, in order to catch singletons generated during the conversion process. We also contract verbal markable spans to their head verb, and merge appositive constructions, which are explicitly marked in GUM, into single mentions, following the OntoNotes guidelines. The order of conversion steps are shown in Table \ref{tab:conversion}.

\begin{table}[t!]
    \centering
    \begin{tabular}{l}
        \hline
        Order of conversion steps \\
        \hline
        \texttt{> Remove bridging \& cataphora} \\
        \texttt{> Contract verbal spans} \\
        \texttt{> Merge appositions} \\
        \texttt{> Remove NN compounding} \\
        \texttt{> Remove copula} \\
        \texttt{> Remove nested entities} \\
        \texttt{> Adjust chains by definiteness} \\
        \texttt{> Remove singletons} \\
        \hline
    \end{tabular}
    \caption{The order of the steps in the dataset conversion from GUM to OntoNotes scheme.}
    \label{tab:conversion}
\end{table}

\subsection{Coreference relations}

\paragraph{Cataphora}
Cataphora encompasses pronominal elements, including demonstratives, e.g.~\textit{those}, which precede an occurrence of a non-pronominal element that occurs within the same utterance and resolves their discourse referent. GUM specifies cataphora in the coreference annotation while OntoNotes only annotates pronominal markables and discards the resolved non-pronominal elements.

Unlike other relations, cataphora `points forward', i.e.~it is resolved by finding a subsequent lexical phrase corresponding to an earlier underspecified pronoun. As in (\ref{eg:cata_gum}), the pronoun \textit{it} (bolded in the box) is resolved by the subsequent markable within the same utterance.

\begin{exe}
\ex \textit{Before:} \cfbox{black}{\textbf{it}} 's good \lfbox[border-style={solid,none,solid,solid},border-color=black]{to be able to do well} \lfbox[border-style={solid,solid,solid,none},border-color=black]{at the World Cup , to be placed} , but \cfbox{black}{it} also means that you get a really good opportunity to know where you 're at in that two year gap between the Paralympics . \label{eg:cata_gum}

\ex \textit{After:} \dbox{it} 's good \lfbox[border-style={solid,none,solid,solid},border-color=black]{to be able to do well} \lfbox[border-style={solid,solid,solid,none},border-color=black]{at the World Cup , to be placed} ,  but \cfbox{black}{it} also means that you get a really good opportunity to know where you 're at in that two year gap between the Paralympics . \label{eg:cata:onto}
\end{exe}   

Because GUM annotates coreference types for each relation, We create a heuristic algorithm to process cataphora. As in (\ref{eg:cata:onto}), the expletive pronoun (dashed box) is removed from the cluster it originally belongs to, leaving it as a singleton. Other coreference relations remain the same in that cluster.

\paragraph{Definiteness}
A definite marker indicates the referent is identifiable in a given context while indefinite markers often function to introduce new entities not mentioned before. Following these tendencies, OntoNotes does not allow coreference relations between an indefinite nominal and any kind of antecedents.
GUM, however, only considers whether the markables refer to the same entity or not. For example, the indefinite noun phrase \textit{a farm several miles outside of town} in (\ref{eg:def_gum}) can occur in the middle of the coreferring chain.

\begin{exe}
\ex \textit{Before:} Rachel Rook took Carroll \cfbox{black}{home} to meet her parents two months after she first slept with him ... her parents lived on \cfbox{black}{a farm several miles outside of town} ... He knew that they never left the \cfbox{black}{farm} ... \label{eg:def_gum}

\ex \textit{After:} Rachel Rook took Carroll \dbox{home} to meet her parents two months after she first slept with him ... her parents lived on \cfbox{black}{a farm several miles outside of town} ... He knew that they never left \cfbox{black}{the farm} ... \label{eg:def_onto}
\end{exe}   

We use the POS tags, dependency labels, and lemmas to distinguish definite and indefinite markables. A definite nominal markable satisfies one of the following cases: (1) it is a pronoun; (2) the head noun is possessed; (3) it is a proper noun; and (4) anything which has the dependency relation \textsc{det} with a definite determiner lemma such as \textit{the}, \textit{that}, etc. The converted document looks like (\ref{eg:def_onto}), where the indefinite nominal phrase and its original antecedent (dashed box) fall in different clusters and the indefinite markable is the first mention in the new cluster\footnote{If the markable \textit{home} also has an antecedent, the coreference relation is not affected by the chain-breaking operation.}.

\subsection{Markables}
\paragraph{Noun-Noun compounding}

In Universal Dependencies (UD), a noun compound is a relation that connects common noun modifiers. GUM marks the noun compounding span in coreference annotations while OntoNotes annotation guidelines do not permit any common noun compound modifiers. As in (\ref{eg:comp_gum}), the two compounds are marked in one coreference chain.

\begin{exe}
\ex \textit{Before:} Allergan Inc. said it received approval to sell the PhacoFlex intraocular lens, the first foldable silicone lens available for \cfbox{black}{cataract surgery} . The lens' foldability enables it to be inserted in smaller incisions than are now possible for \cfbox{black}{cataract surgery} . \label{eg:comp_gum}
\ex \textit{After:} ... available for \dbox{cataract surgery} ... possible for \dbox{cataract surgery} . \label{eg:comp_onto}
\end{exe} 

\noindent To convert the dataset, the conversion program recursively removes the compounding construction from a coreference chain and create a singleton span for that compound, as the two dashed markable spans in (\ref{eg:comp_onto}).

\paragraph{Bridging and split antecedents}
Bridging coreference occurs when two entities do not corefer exactly, but the basis for the identifiability of one referent is the previous mention of one or more previous referents. Particularly, in GUM, an anaphor may have multiple antecedents and each antecedent creates a part-whole relationship with the referent (split antecedent). This part-whole relation, however, is not a valid coreference relation in OntoNotes because two nominal mentions are not semantically identical.

\begin{exe}
\ex \label{eg:bridge_gum} \textit{Before:} \cfbox{black}{Claire Bailey-Ross} ... \cfbox{black}{Andrew Beresford} ... \cfbox{black}{Daniel Smith} ... In this paper, \cfbox{black}{we} report upon the novel insights ... \cfbox{black}{We} will discuss the potential implications ...
\\\\
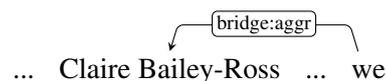
\begin{dependency}
\begin{deptext}[column sep=0.15cm]
... \& Claire Bailey-Ross \& ... \& we \\
\end{deptext}
\depedge[edge unit distance=1ex]{4}{2}{bridge:aggr}
\end{dependency}

\ex \textit{After:} \dbox{Claire Bailey-Ross} ... \dbox{Andrew Beresford} ... \dbox{Daniel Smith} ... In this paper, \cfbox{black}{we} report upon the novel insights ... \cfbox{black}{We} will discuss the potential implications ... \label{eg:bridge_onto}

\end{exe}

As in the example (\ref{eg:bridge_gum}), the three proper nouns link to the collective nominal \textit{we}. With the rich coreference annotation in GUM, all coreference relations identified as \textsc{bridge}, including both split antecedent and other types of bridging anaphora, are removed, possibly leaving the nominals with the original bridging relations as singletons. The example (\ref{eg:bridge_onto}) shows the coreference relations after the conversion process. The proper names are unlinked (dashed boxes) from the original cluster while the pronouns are not affected.

\paragraph{Copula}
Copula predicates are annotated in GUM while are not markables in OntoNotes. If two markables are coreferred, the conversion process utilizes the UD annotations to identify whether or not a copula construction is between the two mention spans. If the \textsc{root} is within the second span and it is the head of \textsc{cop}, we decide to remove the second span from the coreference chain.
\\\\\\\\
\begin{exe}
\ex \textit{Before:} \cfbox{black}{The viewing experience of art} is \cfbox{black}{a complex one} ... The time \cfbox{black}{it} takes ... \label{eg:cop_gum}
\\\\
\begin{dependency}
\begin{deptext}[column sep=0.15cm]
... experience ... \& is \& a \& complex \& one \\
\end{deptext}
\deproot{5}{root}
\depedge[edge unit distance=1ex]{5}{2}{cop}
\depedge[edge unit distance=2ex]{5}{1}{nsubj}
\end{dependency}

\ex \textit{After:} \cfbox{black}{The viewing experience of art} is \dbox{a complex one} ... The time \cfbox{black}{it} takes ... \label{eg:cop_onto}
\end{exe} 

\noindent For example, in (\ref{eg:cop_gum}), the head noun \textit{one} in the second span \textit{a complex one} is the \textsc{root} and the head of the copula \textit{is}. Additionally, it corefers with a markable that is the subject of the copula construction. Therefore, we unlink the second span from the cluster and connect the first span with the markable that the second span refers to. The post-conversion annotations are marked in (\ref{eg:cop_onto}).

\paragraph{Nested entity}
A proper noun may be included in a larger proper noun mention span, as in the nested proper noun \textit{America} within the span \textit{Bank of America}. In OntoNotes, all proper names are considered to be atomic, so that \textit{America} will not be annotated in the above example. Differently from OntoNotes, GUM allows all nested entities to be considered as valid referents. Because a nominal mention is a candidate referent, its possessive modifier, which is also considered as a candidate referent, should be removed from the coreference relation.

\begin{exe}
\ex \textit{Before:} 
I 'm about to go visit and would like to know \lfbox[border-style={solid,none,solid,solid},border-color=black]{the best way to communi-} \lfbox[border-style={solid,solid,solid,none},border-color=black]{cate with her if \cfbox{black}{it} 's helpful} . \label{eg:nmodposs_gum}

\ex \textit{After:} 
I 'm about to go visit and would like to know \lfbox[border-style={solid,none,solid,solid},border-color=black]{the best way to communi-} \lfbox[border-style={solid,solid,solid,none},border-color=black]{cate with her if \dbox{it} 's helpful} . \label{eg:nmodposs_onto}
\end{exe} 

For example, in (\ref{eg:nmodposs_gum}), \textit{it} and \textit{the best way to communicate with her if it's helpful} refer to the same entity, while the pronoun is within the span of the noun phrase, so we remove \textit{it} as in (\ref{eg:nmodposs_onto}). In general, the conversion process removes the nested entity when it is included by its antecedent. 

\paragraph{Singletons}
Singletons are markables that are not referred to by other mentions in a document. GUM explicitly annotates all nominal spans while OntoNotes only considers co-referring noun phrases as valid mention spans. Exemplified in (\ref{eg:sing_gum}), the dashed box indicates that the span is not referred to in the context.

\begin{exe}
\ex \textit{Before:} a unique collection of \dbox{17th Century} Zurbarán paintings \label{eg:sing_gum}
\end{exe} 

The singleton removal process is the last layer for the conversion because previous steps may delete coreference relations and create new markables that are not referred to again.

\subsection{Agreement study}

To evaluate conversion accuracy, three annotators, including an original OntoNotes project member, conducted an agreement study on 3 documents, containing 2,500 tokens and 371 output mentions. Re-annotating from scratch based on OntoNotes guidelines, the conversion achieves a span detection score of $\sim$96 and CoNLL coreference score of $\sim$92, approximately the same as human agreement scores on OntoNotes. After adjudication, the conversion was found to make only 8/371 errors, in addition to 2 errors due to mistakes in the original GUM data, meaning that degradation due to conversion errors is marginal, and consistency should be close to the variability in OntoNotes itself.

\begin{table*}[t!hb]
    \centering\small
    \begin{tabular}{l|cccccccccc|ccc}
    
    \multirow{2}{*}{Genre} & \multicolumn{3}{c}{MUC} & \multicolumn{3}{c}{B$^3$} & \multicolumn{3}{c}{CEAF$_{\phi4}$} && \multicolumn{3}{c}{Mention Detection} \\
     & P & R & F1 & P & R & F1 & P & R & F1 & Avg. F1 & P & R & F1\\
     \hline\hline
    
    & \multicolumn{13}{c}{dcoref}\\\hline
    \textit{academic} & 35.1 & 37.5 & 36.2 & 32.6 & 34.4 & 33.5 & 35.7 & 37.5 & 36.6 & 35.4 & 48.3 & 51.3 & 49.8 \\
    \textit{bio} & 58.0 & 61.6 & 59.8 & 36.8 & 43.6 & 39.9 & 32.1 & 33.5 & 32.8 & 44.1 & 58.9 & 62.3 & 60.6 \\
    \textit{conversation} & 62.2 & 52.9 & 57.1 & 40.5 & 36.7 & 38.5 & 37.1 & 38.2 & 37.6 & 44.4 & 76.6 & 67.8 & 72.0\\
    \textit{fiction} & 57.7 & 43.9 & 49.9 & 50.4 & 33.2 & 40.0 & 37.1 & \textbf{49.0} & \textbf{42.2} & 44.0 & 68.2 & 59.0 & 63.3\\
    \textit{interview} & 57.3 & 53.3 & 55.2 & \underline{29.3} & \underline{21.6} & \underline{24.8} & \underline{22.4} & \underline{24.6} & \underline{23.5} & \underline{27.6} & 64.3 & 60.3 & 62.2\\
    \textit{news} & 57.6 & 55.2 & 56.4 & 45.7 & 42.3 & 44.0 & 39.6 & 32.5 & 35.7 & 45.3 & 44.0 & 50.2 & 46.9 \\
    \textit{reddit} & 59.6 & 65.1 & 62.3 & 38.3 & 53.5 & 44.6 & 32.9 & 34.0 & 33.5 & 46.8 & 60.5 & 64.6 & 62.5 \\
    \textit{speech} & 50.6 & 56.2 & 53.2 & 40.1 & 43.9 & 41.9 & \textbf{46.5} & 38.6 & \textbf{42.2} & 45.8 & 63.5 & 64.2 & 63.9\\
    \textit{textbook} & 36.0 & 34.2 & 35.1 & 32.7 & 31.0 & 31.9 & 23.9 & 39.9 & 29.9 & 32.3 & 18.1 & 45.8 & 26.0 \\
    \textit{vlog} & \textbf{63.6} & \textbf{69.4} & \textbf{66.4} & \textbf{56.4} & \textbf{60.8} & \textbf{58.5} & 31.4 & 36.2 & 33.6 & \textbf{52.8} & 76.4 & 76.8 & 76.6\\
    \textit{voyage} & \underline{34.7} & 37.1 & 35.9 & 30.7 & 28.7 & 29.7 & 29.7 & 35.8 & 32.5 & 32.7 & 46.6 & 62.4 & 53.3\\
    \textit{whow} & 35.8 & \underline{24.2} & \underline{28.9} & 30.0 & 24.5 & 27.0 & 29.9 & 34.0 & 31.8 & 29.2 & 50.0 & 42.9 & 46.2\\

    \hline
    \rule{0pt}{2ex}All OntoGUM & 45.7 & 47.0 & 46.3 & 17.1 & 38.1 & 37.6 & 33.4 & 37.3 & 35.3 & 39.7 & 56.2 & 59.1 & 57.6 \\
    \hline
    \rule{0pt}{2ex}OntoNotes & 57.5 & 61.8 & 59.6 & 68.2 & 68.4 & 68.3 & 47.7 & 43.4 & 45.5 & 57.8 & 66.8 & 75.1 & 70.7 \\
    
    \hline\hline
    
    & \multicolumn{13}{c}{\citet{DBLP:journals/corr/abs-1907-10529}}\\\hline
    \textit{academic} & 84.5 & 53.0 & 65.1 & 83.3 & 48.5 & 61.3 & \textbf{83.2} & 47.0 & 60.1 & 62.2 & 91.0 & 55.2 & 68.7 \\
    \textit{bio} & 85.8 & 74.7 & 79.8 & 61.4 & 64.3 & 62.8 & 65.4 & 49.9 & 56.6 & 66.4 & 87.7 & 74.5 & 80.5\\
    \textit{conversation} & 85.0 & 73.4 & 78.7 & 67.9 & 64.5 & 66.2 & 70.2 & 51.1 & 59.1 & 68.0 & 93.0 & 77.9 & 84.8\\
    \textit{fiction} & \textbf{87.0} & 62.5 & 73.0 & 78.8 & 54.1 & 64.1 & 62.5 & 53.1 & 57.4 & 64.8 & 91.1 & 67.7 & 77.7\\
    \textit{interview} & 83.9 & 71.8 & 77.4 & 76.1 & 60.4 & 67.3 & 72.9 & 50.6 & 59.7 & 68.2 & 85.9 & 70.4 & 77.3\\
    \textit{news} & 65.3 & 65.8 & 65.5 & 60.1 & 59.6 & 59.9 & 58.9 & 54.3 & 56.5 & 60.6 & 71.9 & 70.5 & 71.2 \\
    \textit{reddit} & 76.7 & 67.4 & 71.7 & 67.5 & 60.3 & 63.7 & 69.5 & \underline{40.5} & \underline{51.1} & 61.7 & 85.3 & 68.1 & 75.8 \\
    \textit{speech} & 83.3 & 63.4 & 72.0 & 71.2 & 56.6 & 63.1 & 77.3 & 57.3 & \textbf{65.8} & 67.0 & 91.9 & 69.4 & 79.0\\
    \textit{textbook} & \underline{50.0} & 66.6 & 57.1 & \underline{45.2} & 65.7 & 53.6 & \underline{55.6} & 55.6 & 55.6 & \underline{55.5} & 60.0 & 72.2 & 65.5\\
    \textit{vlog} & \textbf{86.1} & \textbf{86.1} & \textbf{86.1} & 78.4 & \textbf{79.8} & \textbf{79.1} & 63.6 & 47.7 & 54.5 & \textbf{73.3} & 89.4 & 85.4 & 87.4\\
    \textit{voyage} & 69.0 & 70.4 & 69.7 & 52.7 & 64.1 & 57.9 & 65.9 & 53.0 & 58.8 & 62.1 & 78.9 & 75.5 & 77.2\\
    \textit{whow} & 84.8 & \underline{40.9} & \underline{55.2} & \textbf{83.4} & \underline{39.2} & \underline{53.3} & 71.4 & \textbf{57.4} & 63.6 & 57.4 & 93.2 & 52.4 & 67.1\\

    \hline
    \rule{0pt}{2ex}All OntoGUM & 79.7 & 66.3 & 72.4 & 69.5 & 58.58 & 63.7 & 67.7 & 50.7 & 58.0 & 64.6 & 85.4 & 69.2 & 76.5\\
    
    \hline    
    
    \rule{0pt}{2ex}OntoNotes & 85.8 & 84.8 & 85.3 & 78.3 & 77.9 & 78.1 & 76.4 & 74.2 & 75.3 & 79.6 & 89.1 & 86.5 & 87.8\\
    
    \hline
    \end{tabular}
    \caption{Results on the OntoGUM's test dataset with the deterministic coref model (top) and the SOTA coreference model (bottom). The underlined text is the lowest score across 12 genres and bold text is the highest.}
    \label{tab:res}
\end{table*}

\section{Experiments}
We evaluate two systems on the 12 OntoGUM genres, using the official CoNLL-2012 scorer \cite{pradhan-etal-2012-conll, pradhan-etal-2014-scoring}. The primary score is the average F1 of three metrics -- MUC, B$^3$, and CEAF$_{\phi4}$.

\paragraph{Deterministic coreference model}

We first run the deterministic system (dcoref, part of Stanford CoreNLP, \citealt{manning-EtAl:2014:P14-5})
on the OntoGUM benchmark, as it remains a popular option for off-the-shelf coreference resolution. As a rule-based system, it does not require training data, so we directly test it on OntoGUM's test set. However, POS tags, lemmas, and named-entity (NER) information are predicted by CoreNLP, which does have a domain bias favoring newswire. The system's multi-sieve structure and token-level features such as gender and number remain unchanged. We expect that the linguistic rules will function similarly across datasets and genres, notwithstanding biases of the tools providing input features to those rules.

\paragraph{SOTA neural model} Combining the e2e approach with a contextualized LM and span masking is the current SOTA on OntoNotes. The system
utilizes the pretrained SpanBERT-large model, fine-tuned on the OntoNotes training set. Hyperparameters are identical to the evaluation of OntoNotes test to ensure comparable results between the benchmarks. We note that while we choose the SOTA system as a `best case scenario', most off-the-shelf neural NLP toolkits (e.g. spaCy) actually use somewhat simpler e2e models than SpanBERT-large, due to memory/performance constraints.

\section{Results}

\begin{table*}[t!hb]
    \centering\small
    \begin{tabularx}{0.9\textwidth}{  >{\raggedright\arraybackslash}X 
                                    | >{\centering\arraybackslash}X
                                    | >{\centering\arraybackslash}X
                                    | >{\centering\arraybackslash}X
                                    | >{\centering\arraybackslash}X
                                    | >{\centering\arraybackslash}X
                                    }
        Genres & PRON (R) & Other (R) & Total & CoNLL & Span \\\hline
        \textit{vlog} & 600 (.66) & 309 (.34) & 909 & 1 & 1\\
        \textit{interview} & 1223 (.45) & 1485 (.55) & 2708 & 2 & 6\\
        \textit{conversation} & 729 (.61) & 323 (.39) & 1052 & 3 & 2\\
        \textit{speech} & 245 (.40) & 364 (.60) & 609 & 4 & 4\\
        \textit{bio} & 796 (.34) & 1529 (.66) & 2325 & 5 & 3\\
        \textit{fiction} & 1700 (.61) & 1091 (.39) & 2791 & 6 & 5\\
        \textit{academic} & 262 (.21) & 997 (.79) & 1259 & 7 & 10\\
        \textit{voyage} & 300 (.22) & 1053 (.78) & 1353 & 8 & 7\\
        \textit{reddit} & 1337 (.55) & 1077 (.45) & 2414 & 9 & 8\\
        \textit{news} & 340 (.19) & 1483 (.81) & 1823 & 10 & 9\\
        \textit{whow} & 1001 (.47) & 1129 (.53) & 2130 & 11 & 11\\
        \textit{textbook} & 165 (.34) & 315 (.66) & 480 & 12 & 12\\
        \hline
    \end{tabularx}
    \caption{Mention-type counts (ratios) \& ranks of SOTA scores by genre (CoNLL score + span detection).}
    \label{tab:mention_pair}
\end{table*}

\paragraph{OntoGUM vs. OntoNotes} The last rows in each half of Table \ref{tab:res} give overall results for the systems on each benchmark. e2e+SpanBERT encounters a substantial degradation of 15 points (19\%) on OntoGUM, likely due to lower test set lexical and stylistic overlap, including novel mention pairs. We note that its average score of 64.6 is somewhat optimistic, especially given that the system receives access to gold speaker information wherever available (including in \textit{fiction}, \textit{conversation} and \textit{interview}, some of the better scoring genres), which is usually unrealistic. dcoref, assumed to be more stable across genres, also sees losses on OntoGUM of over 18 points (30\%). We believe at least part of the degradation may be due to mention detection, which is trained on different domains for both systems (see the last three columns in the table). These results suggest that input data from CoreNLP degrades substantially on OntoGUM, or that some types of coreferent expressions in OntoGUM are linguistically distinct from those in OntoNotes, or both, making OntoGUM a challenging benchmark for systems developed using OntoNotes.

\paragraph{Comparing genres} Both systems degrade more on specific genres. For example, while \textit{vlog} (with gold speaker information) fares well for both systems, neither does well on \textit{textbook}, and even the SOTA system falls well below (or around) 60s for the \textit{news}, \textit{whow} and \textit{textbook} genres. This might be surprising for \textit{vlog}, which contains transcripts of spontaneous unedited speech from YouTube Creative Commons vlogs quite unlike OntoNotes data; conversely the result is less expected for carefully edited texts which are somewhat similar to data in OntoNotes: OntoNotes contains roughly 30\% newswire text, and it is not immediately clear that GUM's \textit{news} section, which comes from recent Wikinews articles, differs much in genre. Examples (\ref{ex:ultrasounds})--(\ref{ex:report}) illustrate incorrectly predicted coreference chains from both sources and the type of language they contain.

\begin{exe}
\ex \textit{I've been here just crushing ultrasounds ... I've been like crushing \cfbox{black}{these} all day today ... I got sick when I was on Croatia for vacation. I have no idea what it says, but I think \cfbox{black}{they}'re cough drops.} (example from a radiologist's vlog, incorrect: ultrasounds $\neq$ cough drops) \label{ex:ultrasounds}

\ex \textit{\cfbox{black}{The report} has prompted calls for all edible salt to be iodised ... Tasmania was excluded from \lfbox[border-style={solid,none,solid,solid},border-color=black]{the study - where a voluntary iodine for-} \lfbox[border-style={solid,none,solid,none},border-color=black]{tification program using iodised salt in} \lfbox[border-style={solid,solid,solid,none},border-color=black]{bread}, is ongoing} (newswire example, incorrect span and coref: [the study - where a voluntary...]) \label{ex:report}
\end{exe}

These examples show that errors occur readily even in quite characteristic news writing, while genre disparity by itself does not guarantee low performance, as in the case of the vlogs whose lanugage is markedly different. In sum, these observations suggest that accurate coreference for downstream applications cannot be expected in some common well edited genres, despite the prevalence of news data in OntoNotes (albeit specifically from the Wall Street Journal, around 1990). This motivates the use of OntoGUM as a test set for future benchmarking, in order to give the NLP community a realistic idea of the range of performance we may see on contemporary data `in the wild'.

We also suspect that prevalence of pronouns and gold speaker information produce better scores in the results. Table \ref{tab:mention_pair} ranks genres by their e2e CoNLL score, and gives the proportions of pronouns, as well as score rankings for span detection. Because pronouns are usually easier to detect and pair than nouns \citep{durrett-klein-2013-easy}, more pronouns usually means higher scores. On genres with more than 50\% pronouns and gold speakers (\textit{vlog}, \textit{interview}, \textit{conversation}, \textit{speech}, \textit{fiction}) e2e gets much higher results, while genres with few pronouns ($<$30\%) have lower scores (\textit{academic}, \textit{voyage}, \textit{news}). This diversity over 12 genres supports the usefulness of OntoGUM, which can evaluate the genrealizability of coreference systems.

\section{Conclusion}

This paper presented the mechanics of the conversion of the GUM corpus to the OntoGUM version using the OntoNotes coreference scheme, creating the largest open, gold standard coreference dataset following the OntoNotes scheme, which adds several new genres (including more spoken data) to the OntoNotes family. The corpus is automatically converted from GUM by modifying the existing markable spans and coreference relations using multi-layer annotations, such as dependency trees. Results showed a lack of generalizability of existing systems, especially in genres low in pronouns and lacking speaker information. We suspect that at least part of the success of SOTA approaches is due to correct mention detection and high matching scores in genres rich in pronouns, and more so with gold speaker information. Success for other types of mentions in OntoNotes data appears to be much more sensitive to lexical features, performing well on the benchmark test set with high lexical overlap to the training data, but degrading very substantially outside of it, even on newswire texts from our OntoGUM data. This supports use of this challenging dataset for future work, which we hope will benefit evaluations of systems targeting the OntoNotes standard.

\bibliography{anthology,custom}
\bibliographystyle{acl_natbib}

\appendix

\end{document}